\documentclass[11pt,letterpaper]{article}
\usepackage{naaclhlt2016}
\usepackage{times}
\usepackage{latexsym}
\usepackage{amssymb}
\usepackage{amsmath}
\usepackage{graphicx}
\usepackage{url}
\usepackage{changepage}
\usepackage{multirow}
\naaclfinalcopy % Uncomment this line for the final submission
\def\naaclpaperid{***} %  Enter the naacl Paper ID here

\title{DeepStance at SemEval-2016 Task 6: Detecting Stance in Tweets Using Character and Word-Level CNNs}

 \author{Prashanth Vijayaraghavan, Ivan Sysoev, Soroush Vosoughi and Deb Roy \\
         MIT Media Lab, Massachusetts Institute of Technology \\ Cambridge, MA 02139 \\ 
         {\tt pralav@mit.edu}, {\tt isysoev@mit.edu},\\ {\tt soroush@mit.edu}, {\tt dkroy@media.mit.edu}}
\date{}

\begin{document}
\maketitle

\begin{abstract}
%This paper describes our approach for the \emph{Detecting Stance in Tweets} task (SemEval-2016 Task 6). This effort placed us eighth out of 19 teams with macro-average precision, recall and F1-scores of $0.67$, $0.61$ and $0.635$ respectively.
%We utilized recent advances in short text categorization using deep learning to create word-level and character-level models. The choice between word-level and character-level models in each particular case was informed through validation performance. Our final system is a combination of classifiers using word-level or character-level models. We also employed novel data augmentation techniques to expand and diversify our training dataset, thus making our system more robust. 

This paper describes our approach for the \emph{Detecting Stance in Tweets} task (SemEval-2016 Task 6). We utilized recent advances in short text categorization using deep learning to create word-level and character-level models. The choice between word-level and character-level models in each particular case was informed through validation performance. Our final system is a combination of classifiers using word-level or character-level models. We also employed novel data augmentation techniques to expand and diversify our training dataset, thus making our system more robust. Our system achieved a macro-average precision, recall and F1-scores of $0.67$, $0.61$ and $0.635$ respectively.

\end{abstract}

\section{Introduction}
Stance detection is the task of automatically determining whether the authors of a text are against or in favour of a given target. For instance, take the following sentence: "It has been such a cold April, so much for global warming." This sentence's author is most likely against the concept of global warming (i.e., does not believe in it). The work presented here is specifically targeted towards detecting stance in tweets. The noisy and idiosyncratic nature of tweets make this a particularly hard task.

Automatic identification of stance in tweets has practical applications for a range of domains. For instance, it can be used as a sensor to measure the attitude of Twitter users on various issues, such as: political issues, candidates, brand names, TV shows, etc.

There has been extensive research done on modelling and automatic detection of stance in political arenas (e.g., debates) \cite{thomas2006get} and on online forums \cite{somasundaran2009recognizing,murakami2010support}. However, as we alluded to earlier, the peculiar nature of tweets make techniques that have been developed for other platforms unsuitable. The field closest to this work is the field of Twitter sentiment classification, where the task is to detect the sentiment of a given tweet, usually as positive, negative, or neutral. Nonetheless, it is important to note that there are substantial differences between sentiment classification and stance detection. Sentiment classifiers determine the polarity of a given tweet, without considering any targets (see Vosoughi et al. \cite{vosoughi2015enhanced} for an example of a Twitter sentiment classifier). For instance, consider the tweet: "I love Donald Trump", this tweet has a positive sentiment, and the author of the tweet has a positive stance towards Donald Trump, but it can also be inferred that the author is most likely against or at best neutral towards Bernie Sanders.
In this paper, we present a system for automatic detection of stance in Tweets. %The rest of this paper is structured as follows. First, we introduce an overview of our approach a detailed description of our learning models. This is followed by an explanation of our data collection efforts followed by a thorough evaluation of the system. Next, we discuss the performance of our system and end with conclusions from our work and possible paths for future work.

\section{Our Approach}
We trained a different model for each of the five targets. Models for some of the targets used character-level convolutional neural networks(CNN), while other used word-level models. In one particular target (Hillary Clinton), a combination of character-level and word-level models was used. Though Character-level models are robust to the idiosyncratic and noisy nature of tweets, they require a larger dataset compared to word-level models. Our choice between the models was informed by validation performance (as explained in section \ref{eval}). The character and word-level models are explained in the section below.
%Depending on the size and the quality of the data, we employed character-level CNN (CharCNN) model or word-level CNN model (Word-Embedding Convolutional Model) for different targets. 

\subsection{Character-Level CNN Tweet Model}\label{charcnn}
Character-level CNN (CharCNN) is a slight variant of the deep character level convolutional neural network introduced by Zhang et al \cite{zhang2015text}, based on the success of CNNs in image recognition tasks \cite{girshick2014rich} \cite{hinton2012improving}. In this model, we perform temporal (one-dimensional) convolutional and max-pooling operations. Given a discrete input function $f(x)\in [1,l] \mapsto \mathbb{R}$, a discrete kernel function $k(x) \in [1,m] \mapsto \mathbb{R}$ and stride $s$, the convolution $g(y)\in [1,(l-m+1)/s] \mapsto \mathbb{R}$  between $k(x)$ and $f(x)$ and pooling operation $h(y) \in [1,(l-m+1)/s] \mapsto \mathbb{R}$ of $f(x)$ is calculated as:
\begin{equation}
g(y)=\sum_{x=1}^{m}{k(x)\cdot f(y\cdot s-x+c)}
\end{equation}

\begin{equation}
h(y)=max_{x=1}^{m}{f(y\cdot s-x+c)}
\end{equation}
where $c=m-s+1$ is an offset constant. In our implementation of the model, the stride $s$ is set to 1.

\begin{figure}[ht]
  \includegraphics[width=.95\columnwidth]{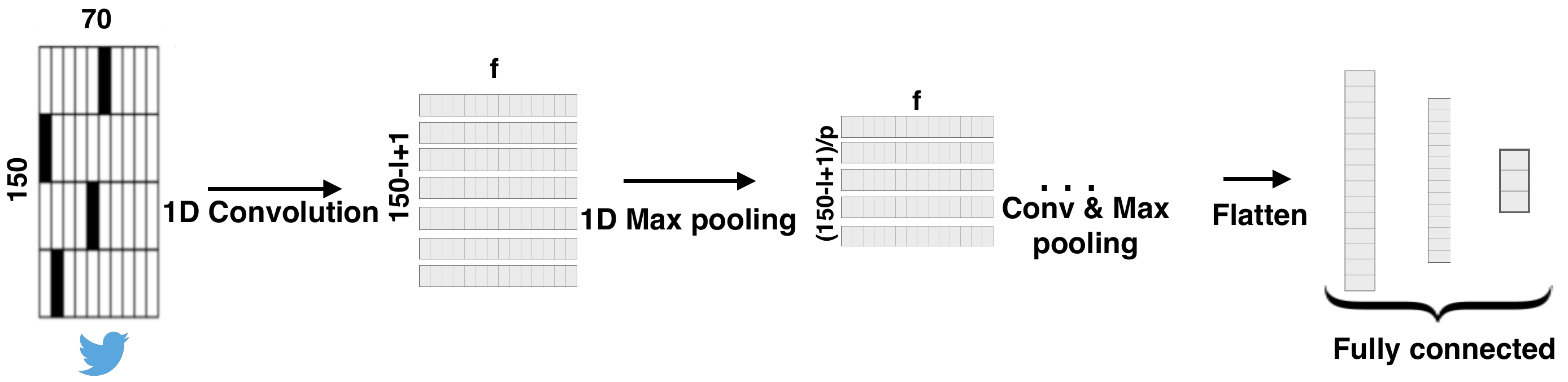}
  \caption{Illustration of CharCNN Model}\label{comp}
\end{figure}

This model is illustrated in Figure \ref{comp}. We adapted this model for the size limit of tweets (140 characters). The character set includes English alphabets, numbers, special characters and unknown character. There are 70 characters in total, given below:\\

 \indent\indent\indent\indent\texttt{abcdefghijklmnopqrstuvw}\\
\indent\indent\indent\indent \texttt{xyz0123456789-,;.!?:'"/}\\
\indent\indent\indent\indent \texttt{\textbackslash|\_\@\#\$\%\&\char`\^ *\textasciitilde`+-=<>()[]\{\}}\\

Each character in the tweet can be encoded using one-hot vector $x_i\in\{0,1\}^{70}$. Hence, a tweet is represented as a binary matrix $x_{1..150}\in\{0,1\}^{150x70}$ with padding wherever necessary, where 150 is the maximum number of characters in a tweet plus padding and 70 is the size of the character set. 
Each tweet, in the form of a matrix, is now fed into a deep model consisting of four 1-d convolutional layers. A convolution operation employs a filter $w$, to extract l-gram character feature from a sliding window of $l$ characters at the first layer and learns abstract textual features in the subsequent layers. This filter $w$ is applied across all possible windows of size $l$ to produce a feature map. A sufficient number ($f$) of such filters are used to model the rich structures in the composition of characters.  Generally, with tweet s, each element $c_i^{(h,F)}(s)$ of a feature map $F$ at the layer $h$ is generated by:

\begin{equation}\label{conv}c_i^{(h,F)}(s)=g(w^{(h,F)}\odot \hat{c}_i^{(h-1)}(s)+b^{(h,F)})\end{equation}
where 
$w^{(h,F)}$ is the filter associated with feature map F at layer $h$;
$\hat{c}_i^{(h-1)}$ denotes the segment of output of layer $h-1$ for convolution at location $i$ (where
$\hat{c}_i^{(0)}=x_{i...i+l-1}$ --- one-hot vectors of $l$ characters from tweet s); $b^{(h,F)}$ is the bias associated with that filter at layer $h$; $g$ is a rectified linear unit and $\odot$ is element-wise multiplication. The output of the convolutional layer $c^h(s)$ is a matrix, the columns of which are feature maps $c^{(h,F_k)}(s)  \vert  k \in 1..f$.

 The output of the convolutional layer is followed by a 1-d max-overtime pooling operation \cite{collobert2011natural} over the feature map and selects the maximum value as the prominent feature from the current filter. Pooling size may vary at each layer (given by $p^{(h)}$ at layer $h$). The pooling operation shrinks the size of the feature representation and filters out trivial features like unnecessary combination of characters (in the initial layer). The window length $l$, number of filters $f$, pooling size $p$ at each layer can vary for each classification task.

The output from the last convolutional layer is flattened and passed into a series of fully connected layers. The output of the final fully connected layer (sigmoid or softmax) gives a probability distribution over categories in our classification task. For regularization we apply a dropout \cite{hinton2012improving} mechanism after the first fully connected layer. This prevents co-adaptation of hidden units by randomly setting a proportion $\rho$ of the hidden units to zero (Generally, we set $\rho=0.5$). CharCNN can be robust to misspellings and noise, provided there is sufficiently large dataset to train the model.
\begin{figure}[ht]
  \includegraphics[width=0.96\columnwidth]{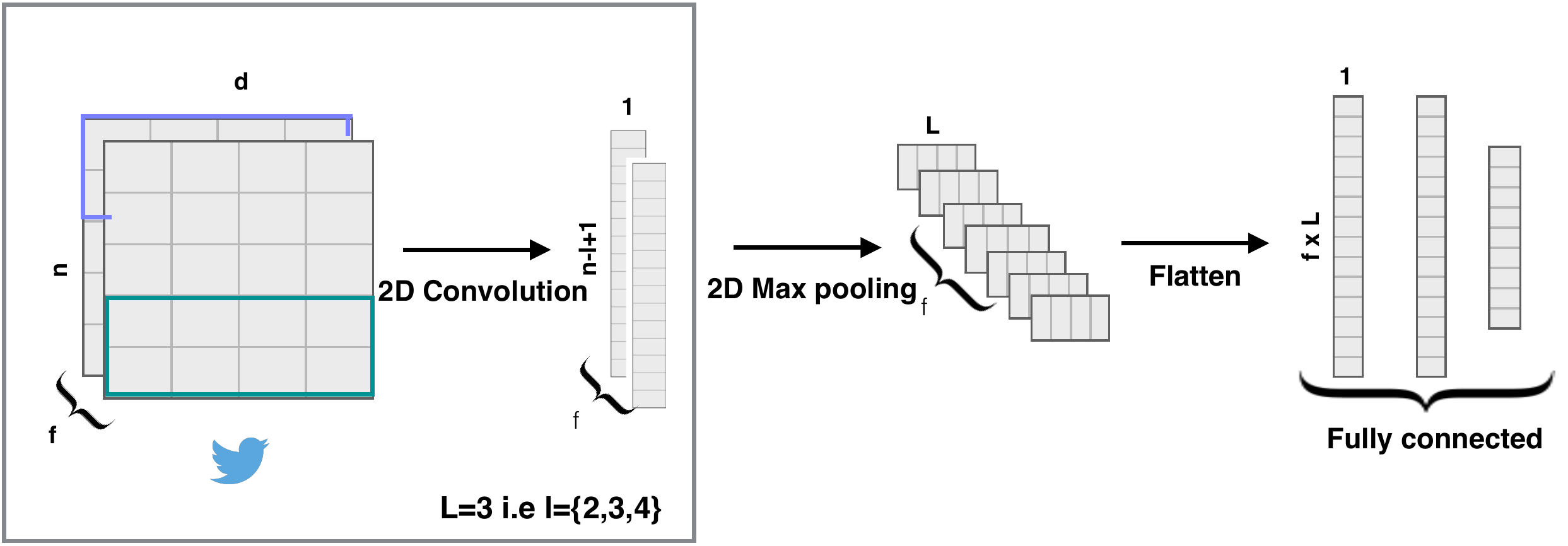}
  \caption{Illustration of Word-Embedding Convolutional Model}\label{word}
\end{figure}
\subsection{Convolutional Word-Embedding Model}\label{wordcnn}
The convolutional embedding model (see Figure \ref{word}) assigns a $d$ dimensional vector to each of the $n$ words of an input tweet resulting in a matrix of size $n \times d$. Each of these vectors are initialized with uniformly distributed random numbers i.e. $x_i \in \mathbb{R}^d$. The model, though randomly initialized, will eventually learn a look-up matrix $\mathbb{R}^{|V|\times d}$ where $|V|$ is the vocabulary size, which represents the word embedding for the words in the vocabulary. 

A convolution layer is then applied to the $n \times d$ input tweet matrix, which takes into consideration all the successive windows of size $l$, sliding over the entire tweet. A filter $w \in \mathbb{R}^{h\times d}$ operates on the tweet to give a feature map $c\in \mathbb{R}^{n-l+1}$. We apply a max-pooling function \cite{collobert2011natural} of size $p=(n-l+1)$ shrinking the size of the resultant matrix by $p$. In this model, we do not have several hierarchical convolutional layers - instead we apply convolution and max-pooling operations with $f$ filters on the input tweet matrix for different window sizes ($l$). 

The vector representations derived from various window sizes can be interpreted as prominent n-gram word features for the tweets. These features are concatenated to give a vector of size $f \times L$, where L is the number of different $l$ values which is further compressed to a size $k$ before passing it to a fully connected softmax or a sigmoid layer whose output is the probability distribution over different categories of our classification task. 

\section{Model Training}
We trained the CharCNN model and the Word-Embedding convolutional model for different targets and selected the best model for each of them. In our task, the tweets are classified into three categories: Favor, Against, and None. We defined the ground truth vector $p$ as a one-hot vector. The commonly used hyperparameters for the convolutional layers of our CharCNN are:  $f=256$, $l=7$ (first two layers) and $l=3$ (other 3 layers). The sizes of the fully connected layers in our CharCNN model are 1,024 and 512.
 
%\begin{table}[h!]%
%\centering

%\begin{tabular}{ |c|c|c|c| }
%\hline %
%$ Layer$ & $Window$  & $Filters$ & $Pool$ $$ \\
%($h$) &$Size$ ($l$) & ($f$)& $Size$ ($p$) \\\hline %
%1 &  7 & 256 & 3 \\\hline
%2 &7 &256 & 3 \\\hline
%3-5&   3 & 256 & N/A\\\hline
%\end{tabular}
%\caption{Commonly used hyperparameters for the Convolutional Layers of CharCNN.}
%\label{charcnnconfig} %
%\end{table}

Similarly, the commonly used hyperparameters of the Convolutional Word-Embedding model are: $l=2,3,4$, $f=200$, $d=300$, $k=256$. Softmax layer takes the output from the penultimate layers of the corresponding models, thereby generating a distribution over the three classes in our task. The class with the maximum probability is the label for the given input tweet.

%\begin{table}[h!]%
%\centering

%\begin{tabular}{ |c|c| }
%\hline %
%$ Hyperparameters$ & $Values$    \\\hline
%Window Size $(l)$ &$2,3,4$  \\\hline
%Filters $(f)$ &$200$ \\\hline
%Embedding Size $(d)$ & $300$ \\\hline
%Penultimate Layer Size $(k)$ & $256$ \\\hline

%\end{tabular}
%\caption{Commonly used hyperparameters for Word-Embedding Convolutional Model}
%\label{wordcnnconfig} %
%\end{table}

To learn the parameters of the model we minimize the cross-entropy loss as the training objective using the Adam Optimization algorithm \cite{kingma2014adam}. It is given by
\begin{equation}\label{cat_ce} CrossEnt(p,q)=-\sum p(x)\log(q(x))\end{equation}
where p is the true distribution (1-of-C representation of ground truth) and q is the output of the softmax. This, in turn, corresponds to computing the negative log-probability of the true class. Each of the classifiers were trained for approximately 8-10 epochs. 

In order to deal with the imbalance in the data, we used a simple balancing technique: to choose a sample on each training step, we randomly picked a class and then randomly selected a tweet associated with this class.

\section{Training Set Expansion}
We expanded the training set by collecting additional tweets for each target-stance pair from the Twitter historical archives. To form a query for the historical API, we automatically selected 40 representative hashtags for each target-stance pair and manually filtered the resulting hashtags lists. The total amount of additional tweets was 1.7 million. Since number of collected tweets vastly exceeded the size of the official dataset, we decided to abstain from using the latter for training and instead use it for validation purposes. For some targets (mentioned in the section \ref{valresults}), we augmented the collected set with tweets obtained by replacing some words and phrases with similar ones, using Word2Vec.

\subsection{Identifying Representative Hashtags}
We found hashtags well-suited for forming a data expansion query. Hashtags are commonly used to represent a ``topic'' or ``theme'' of a tweet and thus often convey information of both the target and the stance (e.g. \#stophillary2016). 

We measured the strength of association between a hashtag and a particular target-stance pair by computing mutual information between them. More precisely, we defined two indicator variables for hashtag occurrences in tweets:
\begin{enumerate}
  \item Whether the current hashtag is equal to the hashtag of interest.
  \item Whether the tweet has the target and the stance of interest.
\end{enumerate}

The mutual information between two random variables is computed as:

\begin{equation} 
I(X, Y) = \sum_{x \in X, y \in Y}p(x,y) log {\frac{p(x,y)}{p(x)p(y)}}
\end{equation}

We estimate mutual information between our indicator variables using a Bayesian approach. We find the expected value of mutual information, assuming an uninformed Dirichlet prior on the joint distribution of the two variables. It can be approximately computed using the formula provided in \cite{hutter2002}:

\begin{equation} 
E\lbrack I\rbrack \approx
 \sum_{i,j \in \lbrace 0, 1\rbrace} \frac{n_{ij}}{n}log{\frac{n_{ij}n}{n_{i+}n_{+j}}} + \frac{0.5}{n}
\end{equation}

Where $n_{ij}$ is the count of samples with indicator variables assuming values $i$ and $j$ respectively, corrected by a pseudo-count of 0.5; $n_{i+} = \sum_j n_{ij}$ and $n_{+j} = \sum_i n_{ij}$.

To get a more reliable estimation of hashtag frequencies for tweets unrelated to the targets, we collected a ``background'' sample of 1.2 million English-language tweets. We treated these tweets as having no stance in relation to any of the targets and used them in computation of the counts above.

For each target-stance pair, we selected 40 hashtags with highest mutual information for further manual filtering. Samples of selected hashtags can be seen in Table 1. The manual filtering step was necessary, since the statistical association with a target-stance pair could only serve as a proxy for the fact that the tag explicitly expresses the target and the stance. For example, \#tcot (standing for ``top conservatives on Twitter'') was highly associated with the stance ``AGAINST'' for the target ``Climate Change is a Real Concern'', but not explicitly expressing this stance. Although we did not make the identification of representative hashtags completely automatic, we found that hashtag filtering is a very manageable task for the annotator, taking only an hour of time for all five targets, making it an ideal place to introduce minimal human input.

\begin{table}[h!]%
\begin{adjustwidth}{0cm}{}
\small
\centering
\begin{tabular}{ | l | c | r | }
\hline
  $Target$ & $FAVOR$ $(F)$ & $AGAINST$ $ (A)$ \\
\hline
  Abn. & \#antichoice & \#prolifeyouth \\
\hline
  Ath. & \#fuckreligion & \#teamjesus \\
\hline
  Cl. Ch. & \#cfcc15 & \#carbontaxscam \\
\hline
  Fem. & \#yesallwomen & \#gamergate \\
\hline
  H. Cl. & \#hillary4women & \#nohillary2016 \\
\hline
\end{tabular}
\caption{Samples of representative hashtags}
\label{hashtags} %

\end{adjustwidth}
\end{table}

\subsection{Collecting and Preprocessing Tweets}

{\renewcommand{\arraystretch}{1.25}
\begin{table}[h!]%
\small
\begin{adjustwidth}{0cm}{}
\centering
\begin{tabular}{ | l | c | r | }
\hline
  $Target$ &$FAVOR$ $(F)$ & $AGAINST$ $ (A)$ \\
\hline
  Abortion & 23,228 & 274,769 \\
\hline
  Atheism & 3,041 & 551,193 \\
\hline
  Climate Change & 355,763 & 60,238 \\
\hline
  Feminism & 124,760 & 178,834 \\
\hline
  Hillary Clinton & 40,060 & 82,294 \\
\hline
\end{tabular}
\caption{Number of collected tweets per target and stance}
\end{adjustwidth}
\label{tweetsnumbers} %
\end{table}

As can be seen in Table 2, the collected tweets are very unevenly distributed between target-stance pairs. There are two causes for this: the uneven distribution of different stances on Twitter in general, and uneven number of representative hashtags that we were able to associate with each target-stance pair. For instance, the pair ``Climate Change is a Global Concern: AGAINST'' was represented by only 15 tweets in the training data that was provided, limiting us to only two representative hashtags. Since our deep learning models require balanced amount of samples, we used the balancing technique described in the previous section.

To eliminate the possibility that resulting classifiers would only learn the hashtags in the query, we removed these hashtags from the majority of the collected tweets, keeping them only in 25\% of the tweets.

\subsection{Augmenting Data Using Word2Vec}
Data augmentation techniques are widely used to enhance generalization of models with respect to input transformations that are known to not affect the output significantly. An example application of data augmentation in NLP can be found in \cite{zhang2015text}, where they used thesaurus-based synonym replacement (WordNet \cite{fellbaum1998wordnet}) to generate additional training samples. We applied the technique used by Zhang et al \cite{zhang2015text} to our task, with the difference that we used Word2Vec \cite{mikolov2013distributed} instead of a thesaurus to find similar words. The underlying intuition was that Word2Vec can provide better coverage for phrases related to our targets.

The algorithm of the data augmentation is as follows. At every step, we randomly selected a tweet from the non-augmented training set. We sampled a number $r$ of words/phrases we would like to replace from a geometric distribution with parameter $p$. We then randomly sampled $r$ words/phrases, that are part of the Word2Vec vocabulary from the current tweet. (if $r$ was larger than number of available words/phrases $n$, we used $r \mod n$.) For each of these words/phrases, we retrieved a list of most similar ones in terms of cosine similarity of Word2Vec vectors. We ordered the list in decreasing order of similarity and truncated it to not include items with similarity less than threshold $t$. We then sampled index $s$ of selected replacement from another geometric distribution with parameter $q$ (again, we used modulo if $s$ was too big). The original words/phrases were then replaced, and the tweet was added to the augmented dataset. The particular values of $p$, $q$ and $t$ were 0.5, 0.5 and 0.25 respectively. Using this method, we generated 500,000 extra tweets for each target-stance pair.

\section{Evaluation}\label{eval}
\begin{table}[b!]%
\small
\begin{adjustwidth}{0cm}{}
\centering
\begin{tabular}{ |c|c|c|c|c| }
\hline %
$ Target$&$St.$ & $Precision$  & $Recall$ & $F1$ $$ \\\hline

Abortion&F& $0.44$  & $0.35$ & $0.39$ \\\hline
Abortion&A& $0.73$  & $0.85$ & $0.79$ \\\hline
Atheism&F &$0.34$ &$0.28$ & $0.31$ \\\hline
Atheism&A &$0.79$ &$0.86$ & $0.82$ \\\hline
Clinton&F &   0.42 & 0.22 & 0.29\\\hline
Clinton&A &   0.64 & 0.90 & 0.75\\\hline
Climate&F &0.80 & 0.73  &0.77 \\\hline
Climate&A &N/A & 0 & N/A\\\hline
Feminism&F & 0.24 & 0.64 & 0.35 \\\hline
Feminism&A & 0.72 & 0.43 & 0.54 \\\hline
\textit{All} &F& 0.44 & 0.35 & 0.39 \\\hline
\textit{All} &A& 0.73 & 0.85 & 0.79 \\\hline
\textit{Macro-Avg} &-& 0.59 & 0.64 & 0.61 \\\hline
\end{tabular}
\end{adjustwidth}
\caption{Baseline performance (Naive Bayes classifiers, test data), St. - Stance}
\label{baseline}
\end{table}
\subsection{Baseline}

\begin{table*}[ht!]%
\centering
\small
\label{devword} %
\begin{tabular}{| p{0.10\linewidth}|p{0.05\linewidth}|p{0.10\linewidth}|p{0.10\linewidth}|p{0.10\linewidth}|p{0.10\linewidth}|p{0.10\linewidth}|p{0.10\linewidth}|} %{ |c|c|c|c|c|c|c|c| }
\hline %
$ Target$ &$St.$&$Precision $ & $Recall$ & $F1 $&$Precision$ & $Recall$ & $F1$    \\
 &&(Word) & (Word) & (Word)&(CharCNN) & (CharCNN) & (CharCNN)    \\\hline
Climate  &A&\textbf{1.00} &$0.27$&$0.42$ &$0.55$ &$\textbf{0.41}$&$\textbf{0.47}$ \\\hline
Climate  &F& $\textbf{0.80}$&$0.67$&$0.73$ & $0.69$&$\textbf{0.80}$&$\textbf{0.74}$   \\\hline
Clinton&A&$0.72$&$\textbf{0.83}$&$\textbf{0.77}$&$\textbf{0.76}$&$0.71$&$0.73$\\\hline
Clinton&F&$\textbf{0.63}$& $0.11$ & $0.18$ &$0.54$& $\textbf{0.46}$ & $\textbf{0.50}$\\\hline
Feminism&A&$0.71$&$0.625$&$0.66$&  \textbf{0.73}  &    \textbf{0.64}  &    \textbf{0.68} \\\hline
Feminism&F&$0.51$&$0.38$&$0.44$& \textbf{0.51}      &\textbf{0.40}      &\textbf{0.45}\\\hline

\end{tabular}
\caption{Performance of Word-Level Classifiers for Climate Change, Hillary Clinton and Feminist Movement.}
\label{wordVsChar}
\end{table*}
To have a better sense of our approach's performance, we compared results against a simple baseline. We built a set of Naive Bayes classifiers using bag-of-word features and optimized their parameters using 20-fold cross validation on original training data. We experimented with different thresholds on word count for a word to be included into vocabulary. We also set up separate thresholds for hashtags and at-mentions. After selecting the most promising values of thresholds, priors and the smoothing parameter, we ran the Naive Bayes classifiers on the test data to obtain results shown in Table \ref{baseline}.

\subsection{Validation Results} \label{valresults}
We trained the models using the collected dataset and validated them on the training set provided for the task. The validation results informed the choice between the word-level and the character-level classifiers for each target. Without Word2Vec augmentation, character-level classifier achieved the best performance only for the target ``Feminist Movement''. When Word2Vec augmentation was introduced, the character-level model achieved the best performance for the target ``Climate Change'' and the stance ''FAVOR'' of the target ''Hillary Clinton''. The word-level model performed better for the targets: ``Legalization of Abortion'', ``Atheism'' and the stance ''AGAINST'' of the target ``Hillary Clinton''. We were able to achieve better average performance for the target ''Hillary Clinton'' by combining character-level and word-level classifiers with a simple heuristic: whenever character-level model predicts ``AGAINST'', use that decision, otherwise resort to the decision of word-level model. 

Table \ref{wordVsChar} compares the performance of the character-level and word-level classifiers for the targets where character-level classifiers yielded an advantage. The macro-average F1 validation score was 0.65.

%\begin{table}[ht]%
%\centering
%\label{devchar} %
%\begin{tabular}{ %p{0.25\linewidth}p{0.25\linewidth}p{0.25\linewidth}p{0.25\linewidth}p{0.25\linewidth}p{0.25\%linewidth}} 
%\hline %
%$ Target$ &$St.$&$Precision$ & $Recall$ & $F1$    \\\hline
%Climate  &A&0.55 &$\textbf{0.41}$&$\textbf{0.47}$  \\\hline
%Climate  &F& $0.69$&$\textbf{0.80}$&$\textbf{0.74}$    \\\hline
%Clinton&A&$\textbf{0.76}$&$0.71$&$0.73$\\\hline
%Clinton&F&$0.54$& $\textbf{0.46}$ & $\textbf{0.50}$ \\\hline
%Feminism&A&  \textbf{0.73}  &    \textbf{0.64}  &    \textbf{0.68}  \\\hline
%Feminism&F& \textbf{0.51}      &\textbf{0.40}      &\textbf{0.45}  \\\hline
%\end{tabular}
%\caption{Performance of Character-Level Classifiers with Word2Vec Data Augmentation for %Climate Change, Hillary Clinton and Feminist Movement}
%\end{table}

\subsection{SemEval Competition Results}
Our model was able to achieve a Macro F-score of 0.6354 (placing us eighth out of 19 teams), while the best performing model had a Macro F-score of 0.6782. Table \ref{resultstest} details results on the test data for each target and stance.

\begin{table}[h!]%
\small
\centering
\label{scores} %
\begin{tabular}{ |c|c|c|c|c| }
%\hline %
%\begin{tabular}{lllll}
%\hline
%                  &  &  &  &  \\ \hline
%\multirow{2}{*}{} &  &  &  &  \\ \cline{2-5} 
%                  &  &  &  &  \\ \hline
%\multirow{2}{*}{} &  &  &  &  \\ \cline{2-5} 
%                  &  &  &  &  \\ \hline

\hline
$ Target$ $ (rank) $ & $St.$ &$P$  & $R$ & $F1$ $$ \\\hline
\multirow{2}{*}{Abn. (1)}   &F& $0.54$  &    $0.67$    &  $0.60$  \\\cline{2-5}
                           &A&  $0.86$   &   $0.54$     &$0.66$ \\\hline
\multirow{2}{*}{Ath. (12)}   &F & $0.33$     & $0.25$     & $0.29$  \\\cline{2-5}
                           &A &$0.82$&      $0.73$&      $0.77 $\\\hline
\multirow{2}{*}{H. Cl (9)}  &F&   $0.37$     & $0.53$      &$0.43$ \\ \cline{2-5} 
                           &A&$ 0.68$   &   $0.66 $& $    0.67$\\\hline
\multirow{2}{*}{Cl. Ch. (12)} &F& $0.80$&$0.82$&$0.81$    \\ \cline{2-5} 
                           & A & N/A &$ 0 $& N/A  \\\hline
\multirow{2}{*}{Fem. (8)}   &F& 0.37      &0.40      &0.38  \\ \cline{2-5} 
                           &A&  0.79  &    0.58  &    0.67  \\\hline
\multirow{2}{*}{All (8)} &F&  0.56      &    0.61  &    0.58  \\ \cline{2-5} 
                           &A&  0.78  &    0.61  &    0.68  \\\hline
\textit{Macro-Avg}&-&  0.67  &    0.61  &    0.635  \\\hline
\end{tabular}
\caption{Results on test data, with rank out of the 19 teams.}
\label{resultstest}
\end{table}

%The performance of various different teams are included in the Table \ref{results}.
%\begin{table}[h!]%
%\centering
%
%\label{results} %
%\begin{tabular}{ |c|c| }
%\hline %
%$Team$ & $F1$    \\\hline
%MITRE &$0.6782$  \\\hline
%pkudblab&	$0.6733$\\\hline
%TakeLab	&$0.6683$\\\hline
%PKULCWM	&$0.6576$\\\hline
%ECNU	&$0.6555$\\\hline
%CU-GWU Perspective	&$0.6360$\\\hline
%IUCL-RF	&$0.6360$\\\hline
%\textbf{DeepStance}	&\textbf{$0.6354$}\\\hline

%\end{tabular}
%\caption{Performance of }
%\end{table}
\section{Discussion and Future Work}
An interesting result of our work was that given enough data, character-level models outperformed word-level models for tweet classification (with a dramatic improvement in case of "Hillary Clinton: FAVOR"). Due to the lack of data, it was necessary to resort to a data augmentation technique to generate sufficient amount and diversity of data for character-level model to show its advantage. 
Another interesting finding from our work is the suitability of word2vec-based substitution as a data augmentation technique. As far as we know, word2vec has not previously been used for data augmentation in this manner.

As can be seen in Table \ref{resultstest}, our system did not perform too well for certain target-stance pairs (e.g., Atheism-Against). We hypothesize that the reason for this is the noise and the limited size of the collected training data. Thus, we believe that the performance of the system can be improved through better data expansion and cleaning techniques.

We see several avenues for future improvements. First, it might be beneficial to use unsupervised pre-training for our models (e.g., using autoencoders for Twitter \cite{vvr_sigir2016}). Second, data cleaning can potentially be improved using bootstrapping. This would entail using our current models (optimized for high precision) to gather cleaner data for the second tier of models. It could be repeated while validation performance improves. Finally, because of the constraints of this SemEval task, we did not manually select hashtags or terms commonly associated with target-stance pairs. Inclusion of such hashtags can potentially boost the quality of the dataset, leading to better performance of our models.  

\bibliography{naaclhlt2016}
\bibliographystyle{naaclhlt2016}

\end{document}